\documentclass{article}
\usepackage{spconf,amsmath,graphicx,hyperref}
\usepackage{amssymb,amsfonts}
\usepackage{graphicx}
\usepackage{textcomp}
\usepackage{xcolor}
\usepackage{url}
\usepackage{graphicx}
\usepackage{mathtools}
\usepackage{algorithm}
\usepackage[noend]{algpseudocode}
\usepackage{pifont}%
\newcommand{\cmark}{\ding{51}}%
\newcommand{\xmark}{\ding{55}}%

\usepackage{etoolbox}
\makeatletter
\apptocmd{\thebibliography}{%
  \footnotesize                        
  \setlength{\itemsep}{0pt}     
  \setlength{\parskip}{0pt}
  \setlength{\parsep}{0pt}
}{}{}
\makeatother
\title{Position-Invariant Fine-Tuning of Speech Enhancement Models with Self-Supervised Speech Representations}
\name{Amit Meghanani, Thomas Hain\thanks{
© 2026 IEEE.  Personal use of this material is permitted.  Permission from IEEE must be obtained for all other uses, in any current or future media, including reprinting/republishing this material for advertising or promotional purposes, creating new collective works, for resale or redistribution to servers or lists, or reuse of any copyrighted component of this work in other works.
}}
\address{Speech and Hearing Research Group \\ School of Computer Science, The University of Sheffield, United Kingdom\\
\{ameghanani1,t.hain\}@sheffield.ac.uk}
\begin{document}
\ninept
\maketitle
\begin{abstract}
Integrating front-end speech enhancement (SE) models with self-supervised learning (SSL)-based speech models is effective for downstream tasks in noisy conditions. SE models are commonly fine-tuned using SSL representations with mean squared error (MSE) loss between enhanced and clean speech. However, MSE is prone to exploiting positional embeddings in SSL models, allowing the objective to be minimised through positional correlations instead of content-related information. This work frames the problem as a general limitation of self-supervised representation fine-tuning and investigates it through representation-guided SE. Two strategies are considered: (1) zero-padding, previously explored in SSL pre-training but here examined in the fine-tuning setting, and (2) speed perturbations with a soft-DTW loss. Experiments show that the soft-DTW-based approach achieves faster convergence and improved downstream performance, underscoring the importance of position-invariant fine-tuning in SSL-based speech modelling.
\end{abstract}

\section{Introduction}

Recent advances in deep learning have substantially transformed speech processing technologies, with self-supervised learning (SSL) emerging as a powerful framework for learning from unlabelled data. SSL-based speech models such as wav2vec 2.0 \cite{wav2vec2}, HuBERT \cite{hubert}, and WavLM \cite{wavlm} have demonstrated impressive performance across a variety of downstream tasks including automatic speech recognition (ASR), phoneme recognition (PR), speaker identification (SID), emotion recognition (ER), and learning discriminative acoustic word embeddings (AWE) \cite{hubert,wavlm,ssl_review,layerwise_analysis,asru,eacl}. These models are pre-trained on large amounts of unlabelled speech and can then be fine-tuned on task-specific labelled datasets, substantially reducing the reliance on expensive annotated data \cite{ssl_review}.
Despite their success, SSL models face challenges in noisy environments. Several approaches have been proposed to improve noise robustness through self-supervised fine-tuning \cite{spin}. For example, R-SPIN \cite{rspin} (robust speaker-invariant clustering) enhances robustness by learning discrete acoustic units with speaker-invariant clustering, making the resulting representations both speaker- and noise-invariant. Another recent work, deHuBERT \cite{dehubert}, introduced noise-agnostic representations derived from HuBERT using noise augmentation and a Barlow Twins-based loss \cite{barlow_twin}. WavLM incorporated a denoising task directly into pre-training. Further, \cite{se_integration} demonstrated that integrating a single-channel speech enhancement (SE) frontend improves performance in noisy conditions. However, the SE frontend in that work required joint fine-tuning with a downstream ASR model, limiting its generality and reusability.

To address this limitation, subsequent work proposed the SSL-MSE loss \cite{se_ssl}, which compares enhanced and clean signals in the SSL feature space using mean squared error (MSE). This enables task-agnostic fine-tuning of SE models by aligning their outputs with SSL encoders, independent of downstream task objectives. However, MSE introduces a critical issue: it tends to exploit positional embeddings in SSL models, minimising the loss through positional correlation rather than meaningful content. When applied between the SSL representations of enhanced and clean speech, this behaviour reduces generalisation. The phenomenon is closely related to ``positional collapse'' observed in SSL pre-training \cite{spiral,representation_collapse}, where models reduce loss by exploiting position rather than learning semantically relevant representations. SPIRAL \cite{spiral} addressed this in pre-training by introducing random zero-padding to disrupt positional alignment, thereby encouraging models to focus on content.

The present study investigates the exploitation of positional embeddings in SSL-guided SE fine-tuning and introduces two mitigation strategies. The first, \textit{positional perturbation through random zero-padding}, was introduced in SPIRAL for pre-training; here, it is examined in the fine-tuning setting, providing empirical validation in a new context. The second, \textit{speed perturbation combined with a soft-DTW loss}, ensures temporal alignment between enhanced and clean signals of varying length while reducing reliance on absolute positional encodings. Soft-DTW \cite{soft-dtw,score,laser}, a differentiable version of dynamic time warping (DTW), provides a principled way to achieve content-based alignment.
To evaluate these strategies, SE models are fine-tuned using different loss functions and perturbations, followed by supervised fine-tuning of SSL models with SE frontends on downstream tasks including ASR and PR. Experiments are conducted on a noise-augmented version of LibriSpeech \cite{librispeech} consistent with the SUPERB benchmark \cite{superb}, using environmental recordings from DEMAND \cite{demand} to simulate noisy conditions.

The main contributions of this study are as follows:
\begin{itemize}
    \item Identification and verification of the exploitation of positional embeddings in SSL-MSE-based SE fine-tuning, showing its adverse effect on robustness and generalisation of SE frontends.
    \item Evaluation of two strategies to mitigate this issue: (1) positional perturbation via random zero-padding, previously proposed in pre-training but not validated in fine-tuning, and (2) speech perturbation with soft-DTW alignment loss, which improves performance and convergence speed over the SSL-MSE baseline.
\end{itemize}

The remainder of this paper is organised as follows: Sec.~\ref{sec2} introduces the baseline system and the proposed mitigation strategies; Sec.~\ref{sec3} describes the experimental setup; Sec.~\ref{sec4} presents results and discussion; and Sec.~\ref{sec5} concludes the work.

\section{Methodology}
\label{sec2}

\begin{figure}
    \centering
    \includegraphics[width=0.9\columnwidth]{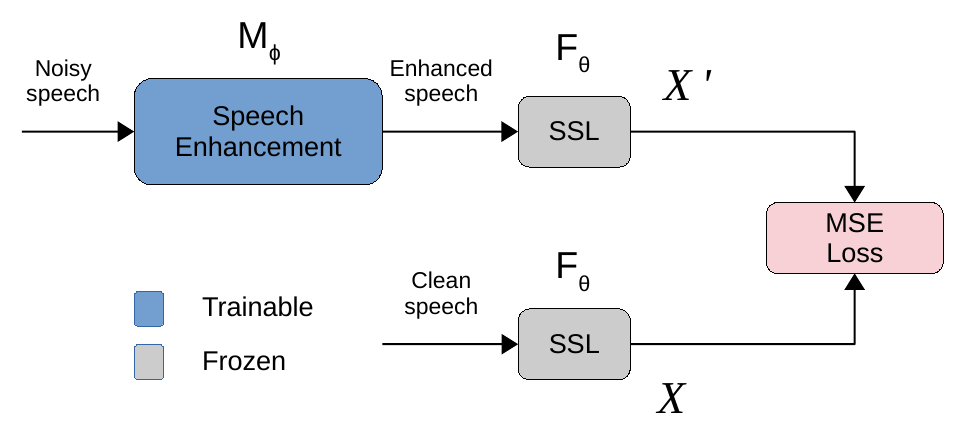}
    \caption{SSL-MSE: pipeline for fine-tuning a frontend SE model using SSL-based speech representations with MSE loss \cite{se_ssl}.}
    \label{fig1}
\end{figure}

\subsection{Baseline}
\label{sec2a}
The pipeline for fine-tuning the speech enhancement model with SSL-based speech representations \cite{se_ssl} is shown in Fig.~\ref{fig1}. A noise-augmented version of a speech signal is passed to the SE model, which outputs enhanced speech. Then, SSL representations are extracted from the SSL model for both the enhanced signal and the original clean speech signal. These representations are compared using mean squared error (MSE) loss. Let $X = \{x_1, x_2, \dots, x_m\}$ and $X' = \{x_1', x_2', \dots, x_m'\}$ be the \( d \)-dimensional representations obtained from the clean and enhanced speech, respectively, from a particular layer of the SSL model. Then, for a single training sample, the SSL-MSE loss is defined as:
\begin{equation}
\mathcal{L}_{\text{SSL-MSE}}(X',X) = \frac{1}{m} \sum_{i=1}^{m} \left\| x_i - x_i' \right\|_2^2
\end{equation}

As discussed in the introduction, MSE is prone to exploiting positional embeddings present in SSL models. This behaviour allows the loss to minimise itself through absolute positional correlation rather than meaningful content similarity, which limits generalisation. Two methods to mitigate this issue are described below.

\subsection{Positional Perturbation through Random Zero-Padding}
\label{sec2b}

\begin{algorithm}[t!]
\footnotesize
\caption{Fine-tuning SE model with positional perturbation via random zero-padding (SSL-MSE-PAD)}
\begin{algorithmic}[1]
\State $M_{\phi}$ = Pre-trained SE model
\State $F_{\theta}$ = Frozen SSL model
\State Total samples in dataset = $N_{samp}$
\State $S_i = (\mathbf{s}_{\text{clean}}, \mathbf{s}_{\text{noisy}})$ = i\textsuperscript{th} clean-noisy pair
\While{Not Converged}
  \For{\texttt{i = 1 to $N_{samp}$}}
    \State Randomly sample $p \in [0.02, 0.05]$
    \State $L_p = \left\lfloor \frac{p \cdot T}{320} \right\rfloor \cdot 320$
    \State $\mathbf{s}_{\text{clean}}^{\text{pad}} = [\mathbf{0}_{L_p}, \mathbf{s}_{\text{clean}}, \mathbf{0}_{L_p}]$
    \State $\mathbf{s}_{\text{enh}} = M_{\phi}(\mathbf{s}_{\text{noisy}})$
    \State $X = F_{\theta}(\mathbf{s}_{\text{clean}}^{\text{pad}}) $, $X  \in \mathbb{R}^{n \times d}$
    \State $X' = F_{\theta}(\mathbf{s}_{\text{enh}})$, $X'  \in \mathbb{R}^{m \times d}$
    \State $r = L_p / 320$
    \State $\hat{X} = \{x_{r+1}, x_{r+2}, \dots, x_{n-r}\}$, $\hat{X}  \in \mathbb{R}^{m \times d}$
    \State $\mathcal{L_{\text{SSL-MSE-PAD}}} = \text{MSE}(X', \hat{X})$
    \State Compute gradients $\frac{\partial \mathcal{L_{\text{SSL-MSE-PAD}}}}{\partial \phi}$
    \State Update $\phi$ to minimize $\mathcal{L_{\text{SSL-MSE-PAD}}}$
  \EndFor
\EndWhile
\end{algorithmic}
\label{alg1}
\end{algorithm}

To discourage reliance on absolute positional information, a random zero-padding strategy is applied to the clean reference waveform. This idea was first proposed in the SPIRAL framework \cite{spiral} for SSL pre-training, and is here evaluated in the fine-tuning context.  
Let the clean waveform be represented by \( \mathbf{s}_{\text{clean}} \in \mathbb{R}^{T} \), where \( T \) is the number of samples. For each training sample, a padding percentage \( p \in [0.02, 0.05] \) is randomly chosen, and the padding length is computed as:
\[
L_p = \left\lfloor \frac{p \cdot T}{320} \right\rfloor \cdot 320,
\]
ensuring consistency with the SSL model’s frame hop size (20 ms at 16 kHz sampling).  

The padded signal is then constructed as:
\[
\mathbf{s}_{\text{clean}}^{\text{pad}} = [\mathbf{0}_{L_p}, \mathbf{s}_{\text{clean}}, \mathbf{0}_{L_p}],
\]
where \( \mathbf{0}_{L_p} \) is a vector of zeros of length \( L_p \).  

Let \( X = \{x_1, x_2, \dots, x_n\} \) denote the $d$-dimensional frame-level representations extracted from the padded clean waveform using an SSL model such as HuBERT. Since the enhanced waveform remains unpadded, \( r = L_p / 320 \) frames are trimmed from both ends of \( X \) to restore alignment:
\[
\hat{X} = \{x_{r+1}, x_{r+2}, \dots, x_{n-r}\}.
\]

This yields \( \hat{X} \in \mathbb{R}^{m \times d} \), matching the dimension of $X' \in \mathbb{R}^{m \times d}$. The SSL-MSE-PAD loss is then:
\begin{equation}
\mathcal{L}_{\text{SSL-MSE-PAD}}(X',\hat{X}) = \frac{1}{m} \sum_{i=1}^{m} \left\| \hat{x}_i - x_i' \right\|_2^2
\end{equation}
The SE model fine-tuning process with SSL-MSE-PAD loss is summarised in Algorithm~\ref{alg1}.
\subsection{Speed Perturbation with Soft-DTW Loss}
\label{sec2c}

\begin{algorithm}[t!]
\footnotesize
\caption{Fine-tuning SE model with speed perturbation and soft-DTW alignment (SSL-SoftDTW)}
\begin{algorithmic}[1]
\State $M_{\phi}$ = Pre-trained SE model
\State $F_{\theta}$ = Frozen SSL model
\State Total samples in dataset = $N_{samp}$
\State $S_i = (\mathbf{s}_{\text{clean}}, \mathbf{s}_{\text{noisy}})$ = i\textsuperscript{th} clean-noisy pair
\While{Not Converged}
  \For{\texttt{i = 1 to $N_{samp}$}}
    \State Sample speed factor $\alpha$
    \State $\mathbf{s}_{\text{pert}} = \text{SpeedPerturb}(\mathbf{s}_{\text{clean}}, \alpha)$
    \State $\mathbf{s}_{\text{enh}} = M_{\phi}(\mathbf{s}_{\text{noisy}})$
    \State $\hat{X} = F_{\theta}(\mathbf{s}_{\text{pert}})$, $\hat{X}  \in \mathbb{R}^{n \times d}$
    \State $X' = F_{\theta}(\mathbf{s}_{\text{enh}})$, $X'  \in \mathbb{R}^{m \times d}$
    \State $\mathcal{L_{\text{SSL-SoftDTW}}} = \frac{\text{soft-DTW}_{\gamma}(X', \hat{X})}{m + n}$
    \State Compute gradients $\frac{\partial \mathcal{L_{\text{SSL-SoftDTW}}}}{\partial \phi}$
    \State Update $\phi$ to minimize $\mathcal{L_{\text{SSL-SoftDTW}}}$
  \EndFor
\EndWhile
\end{algorithmic}
\label{alg2}
\end{algorithm}

This approach combines speed perturbation with soft-DTW-based loss between clean and enhanced representations. Unlike zero-padding, it introduces continuous and local temporal distortions, simulating realistic variability in speech timing while enforcing content-based alignment.  
Let the clean waveform be denoted by \( \mathbf{s}_{\text{clean}} \in \mathbb{R}^T \). A speed perturbation factor \( \alpha \) is sampled uniformly at random to generate:
\[
\mathbf{s}_{\text{pert}} = \text{SpeedPerturb}(\mathbf{s}_{\text{clean}}, \alpha),
\]
where \( \mathbf{s}_{\text{pert}} \in \mathbb{R}^{T'} \) with \( T' \neq T \) when $\alpha \neq 1$. Torchaudio \cite{torchaudio} is used for speed perturbation. 
Let \( \hat{X} = \{\hat{x}_1, \dots, \hat{x}_{n}\} \) be the SSL representations of the perturbed clean waveform, and \( X' = \{x_1', \dots, x_m'\} \) those of the enhanced waveform. Since the two sequences differ in length and temporal structure, direct framewise MSE cannot be applied. Instead, alignment is achieved using soft dynamic time warping (soft-DTW) \cite{soft-dtw}, a differentiable extension of DTW that replaces the hard minimum operator with a soft-min (log-sum-exp) function.  

The SSL-SoftDTW loss is defined as:
\begin{equation}
  \mathcal{L}_{\text{SSL-SoftDTW}}(X', \hat{X}) =  \frac{\text{soft-DTW}_{\gamma}(X', \hat{X})}{m + n}  
\end{equation}

Normalisation by $m+n$ compensates for sequence length dependence. A smoothing factor $\gamma = 0.1$ is used, following common practice \cite{soft-dtw}. To handle potential negative values, a divergence-based normalisation of soft-DTW is employed \cite{diff_divergence,tslearn,lav}.  

The SE model fine-tuning process with SSL-SoftDTW loss is summarised in Algorithm~\ref{alg2}.

\section{Experimental Setup}
\label{sec3}

\subsection{Frontend SE Model Fine-tuning}
\paragraph*{SE Backbone:}
The speech enhancement backbone used in this study is the \texttt{master64} model from Facebook Research’s \sloppy\texttt{Denoiser} toolkit\footnote{\label{fn:denoiser}\url{https://github.com/facebookresearch/denoiser}}, which contains 33.5 million parameters. \texttt{master64} is a deep convolutional time-domain speech enhancement network designed for real-time operation with low latency \cite{se_model}. Architecturally, it is based on a modified version of the Demucs (Deep Extractor for Music Sources) network, featuring multiple convolutional encoder and decoder layers with skip connections and long short-term memory (LSTM) blocks in the bottleneck. The \texttt{master64} variant is notable for its relatively small receptive field and reduced model size, using 64 feature channels throughout. Unlike spectral masking approaches, \texttt{master64} directly operates on raw audio waveforms, learning to map noisy speech to its clean counterpart. The model is pre-trained on large synthetic datasets with diverse noise types and speech sources, and is further fine-tuned in this work as described in Fig.~\ref{fig1} and Sec.~\ref{sec2}. The model is implemented in PyTorch and publicly available as part of the Denoiser repository. 

\paragraph*{Dataset for SE Fine-tuning:}
\label{3c}
For fine-tuning the SE model, a noise-augmented version of the LibriSpeech \cite{librispeech} \texttt{train-clean-100} subset was created. Noise samples were drawn from the DEMAND dataset \cite{demand}, which includes six noise categories grouped into two broad classes: \emph{indoor} (Domestic, Office, Public, Transportation) and \emph{outdoor} (Street, Nature). Only indoor recordings were used for SE fine-tuning. From the 16 available channels in each DEMAND recording, the first channel was used. All noise recordings were downsampled from 48~kHz to 16~kHz. For each utterance, a noise segment was randomly selected and added at a signal-to-noise ratio (SNR) chosen from $\{0, 5, 10, 20\}$~dB.

\paragraph*{Fine-tuning Details:}
The SE system is fine-tuned using the clean and noise-augmented LibriSpeech \texttt{train-clean-100} dataset, as shown in Fig.~\ref{fig1}. In addition to the conventional SSL-MSE loss, two mitigation strategies addressing the exploitation of positional embeddings are considered: (1) positional perturbation via random zero-padding (SSL-MSE-PAD) and (2) speed perturbation with soft-DTW alignment loss (SSL-SoftDTW), described in Sec.~\ref{sec2b} and \ref{sec2c}. 
For fine-tuning, the BASE version of HuBERT is used as the SSL model, containing roughly 95 million parameters. HuBERT-BASE consists of multi-layer CNNs at the frontend, followed by 12 Transformer layers. The output from the final Transformer layer is a 768-dimensional sequence of vectors. Prior work \cite{se_ssl} reported that using the last layer for SSL-MSE loss performs best across different downstream tasks. Therefore, in this study, representations from the final HuBERT layer are used consistently across all fine-tuning strategies (SSL-MSE, SSL-MSE-PAD, and SSL-SoftDTW). 
The \texttt{master64} SE model was fine-tuned using the Adam optimizer with a learning rate of $1.0 \times 10^{-4}$. Training was performed for one epoch with an effective batch size of 16, achieved via gradient accumulation. During each step, the noisy speech was enhanced by the SE model and then passed through a frozen HuBERT model to extract SSL representations. These were aligned with clean speech representations as discussed in Sec.~\ref{sec2}. All representations were L2-normalized prior to loss computation. Gradient clipping (max-norm = 1.0) was applied at each step.

\subsection{Evaluation on SUPERB Downstream Tasks}
For ASR and PR evaluation, noise-augmented versions of the LibriSpeech \texttt{train-clean-100}, \texttt{dev-clean}, and \texttt{test-clean} subsets used in SUPERB \cite{superb} were created. DEMAND \cite{demand} noises were added at SNRs of $\{0, 5, 10, 20\}$~dB. Indoor noises were used for training and development (seen noise), while evaluation included both indoor and outdoor noises (unseen noise).  
After SE frontend fine-tuning, the HuBERT model augmented with the SE frontend was assessed on ASR and PR using the S3PRL toolkit\footnote{\scriptsize{\url{https://github.com/s3prl/s3prl}}} \cite{distill_HuBERT,tera}. Each task employed a task-specific head with a learnable weighted average of HuBERT layer-wise representations, fine-tuned jointly with the head. For ASR, the head was a two-layer bidirectional LSTM (1024 units per layer) trained with character-level CTC loss \cite{superb,ctc}. Training used \texttt{train-clean-100 + indoor noise} and validation used \texttt{dev-clean + indoor noise}. Evaluation was conducted on \texttt{test-clean}, \texttt{test-clean + indoor noise}, and \texttt{test-clean + outdoor noise}, without external language models. For PR, the head was a linear frame-level classifier trained with CTC loss, following SUPERB \cite{superb}.  
ASR and PR tasks were optimized using Adam with learning rates of $1.0 \times 10^{-4}$ and $5.0 \times 10^{-4}$, respectively. Each configuration was repeated \textbf{five times}, and results are reported as the \textbf{mean and standard deviation} of the \textit{word error rate (WER, \%)} for ASR and the \textit{phoneme error rate (PER, \%)} for PR.

\section{Results and Discussion}
\label{sec4}

\begin{table}[htbp]
\centering
\resizebox{\columnwidth}{!}{%
\setlength{\tabcolsep}{2pt}
\begin{tabular}{ccccc}
\hline
\multicolumn{1}{c|}{\textbf{\begin{tabular}[c]{@{}c@{}}SSL Fine-tuning\\ of SE Model\end{tabular}}} & \multicolumn{1}{c|}{\textbf{Enhancement}} & \multicolumn{1}{c|}{\textbf{\begin{tabular}[c]{@{}c@{}}test-clean \\ +\\  outdoor noise\end{tabular}}} & \multicolumn{1}{c|}{\textbf{\begin{tabular}[c]{@{}c@{}}test-clean\\ +\\ indoor noise\end{tabular}}} & \textbf{test-clean} \\ \hline
\xmark                                                                                                 & \xmark                                        & 12.47 $\pm$ 0.07                                                                                                 & 13.09  $\pm$ 0.07                                                                                             & 6.29   $\pm$ 0.04             \\
\xmark                                                                                                 & \cmark                                       & 9.93 $\pm$ 0.08                                                                                                  & 9.96  $\pm$   0.04                                                                                            & 6.22    $\pm$ 0.05            \\
SSL-MSE                                                                                            & \cmark                                       & 9.19     $\pm$   0.05                                                                                           & 8.89  $\pm$ 0.07                                                                                              & 6.21     $\pm$ 0.08           \\
SSL-MSE-PAD                                                                                        & \cmark                                       & 9.11  $\pm$ 0.05                                                                                                 & 8.86  $\pm$ 0.03                                                                                              & 6.21  $\pm$ 0.04              \\
SSL-SoftDTW                                                                                            & \cmark                                       & \textbf{9.06}  $\pm$ 0.09                                                                                        & 8.88   $0\pm$ 0.08                                                                                    & 6.21 $\pm$ 0.11      \\ \hline
\end{tabular}%
}
\caption{Performance of HuBERT on the ASR task under three test conditions: \texttt{test-clean}, \texttt{test-clean + indoor noise}, and \texttt{test-clean + outdoor noise}. The ASR model head is fine-tuned using noise-augmented \texttt{train-clean-100} as the training set and \texttt{dev-clean} as the validation set.
}

\label{tab1}
\end{table}
\begin{table}[htbp]
\centering
\resizebox{\columnwidth}{!}{%
\setlength{\tabcolsep}{2pt}
\begin{tabular}{ccccc}
\hline
\multicolumn{1}{c|}{\textbf{\begin{tabular}[c]{@{}c@{}}SSL Fine-tuning\\ of SE Model\end{tabular}}} & \multicolumn{1}{c|}{\textbf{Enhancement}} & \multicolumn{1}{c|}{\textbf{\begin{tabular}[c]{@{}c@{}}test-clean \\ +\\  outdoor noise\end{tabular}}} & \multicolumn{1}{c|}{\textbf{\begin{tabular}[c]{@{}c@{}}test-clean\\ +\\ indoor noise\end{tabular}}} & \textbf{test-clean} \\ \hline
\xmark                                                                                                 & \xmark                                        & 9.82 $\pm$ 0.03                                                                                                  & 10.36  $\pm$ 0.02                                                                                             & 5.55 $\pm$ 0.04                \\
\xmark                                                                                                 & \cmark                                       & 7.48 $\pm$ 0.03                                                                                                   & 7.55 $\pm$ 0.03                                                                                                & 5.28 $\pm$ 0.02                \\
SSL-MSE                                                                                            & \cmark                                       & 6.78  $\pm$ 0.01                                                                                                  & 6.63 $\pm$ 0.02                                                                                                & 5.17  $\pm$ 0.02              \\
SSL-MSE-PAD                                                                                        & \cmark                                       & 6.85  $\pm$ 0.03                                                                                                 & 6.64    $\pm$ 0.07                                                                                            & 5.18   $\pm$ 0.03             \\
SSL-SoftDTW                                                                                            & \cmark                                       & \textbf{6.70} $\pm$ 0.00                                                                                         & 6.61 $\pm$ 0.02                                                                                      & 5.11 $\pm$ 0.00      \\ \hline
\end{tabular}%
}
\caption{Performance of HuBERT on the PR task under three test conditions: \texttt{test-clean}, \texttt{test-clean + indoor noise}, and \texttt{test-clean + outdoor noise}. The PR model head is fine-tuned using noise-augmented \texttt{train-clean-100} as the training set and \texttt{dev-clean} as the validation set. }

\label{tab3}
\end{table}
\paragraph*{Automatic Speech Recognition:}
Table \ref{tab1} presents WER results for the ASR task of SUPERB under three test sets: \texttt{test-clean}, \texttt{test-clean + indoor noise} (seen noise), and \texttt{test-clean + outdoor noise} (unseen noise). The first row corresponds to the no-enhancement condition, where noisy speech is directly used for training and evaluation. The second row uses the off-the-shelf SE model \texttt{master64}\footref{fn:denoiser}, without any fine-tuning. In this case, speech is enhanced prior to ASR training and evaluation. As expected, incorporating an SE frontend substantially improves performance across all test conditions, consistent with prior findings \cite{se_ssl,se_integration}. 
The next three rows correspond to the fine-tuning strategies described in Sec.~\ref{sec2}. The baseline SSL-MSE approach \cite{se_ssl} is compared with two variants designed to mitigate the exploitation of positional embeddings: SSL-MSE-PAD and SSL-SoftDTW. Among these, SSL-SoftDTW yields the best performance in the unseen noise condition, while SSL-MSE-PAD achieves only marginal gains over the baseline. The limited improvement of SSL-MSE-PAD may be attributed to the introduction of artificial discontinuities from zero-padding, which can interfere with SSL feature extraction.  

\begin{figure}[htbp]
    \centering
    \includegraphics[width=\linewidth]{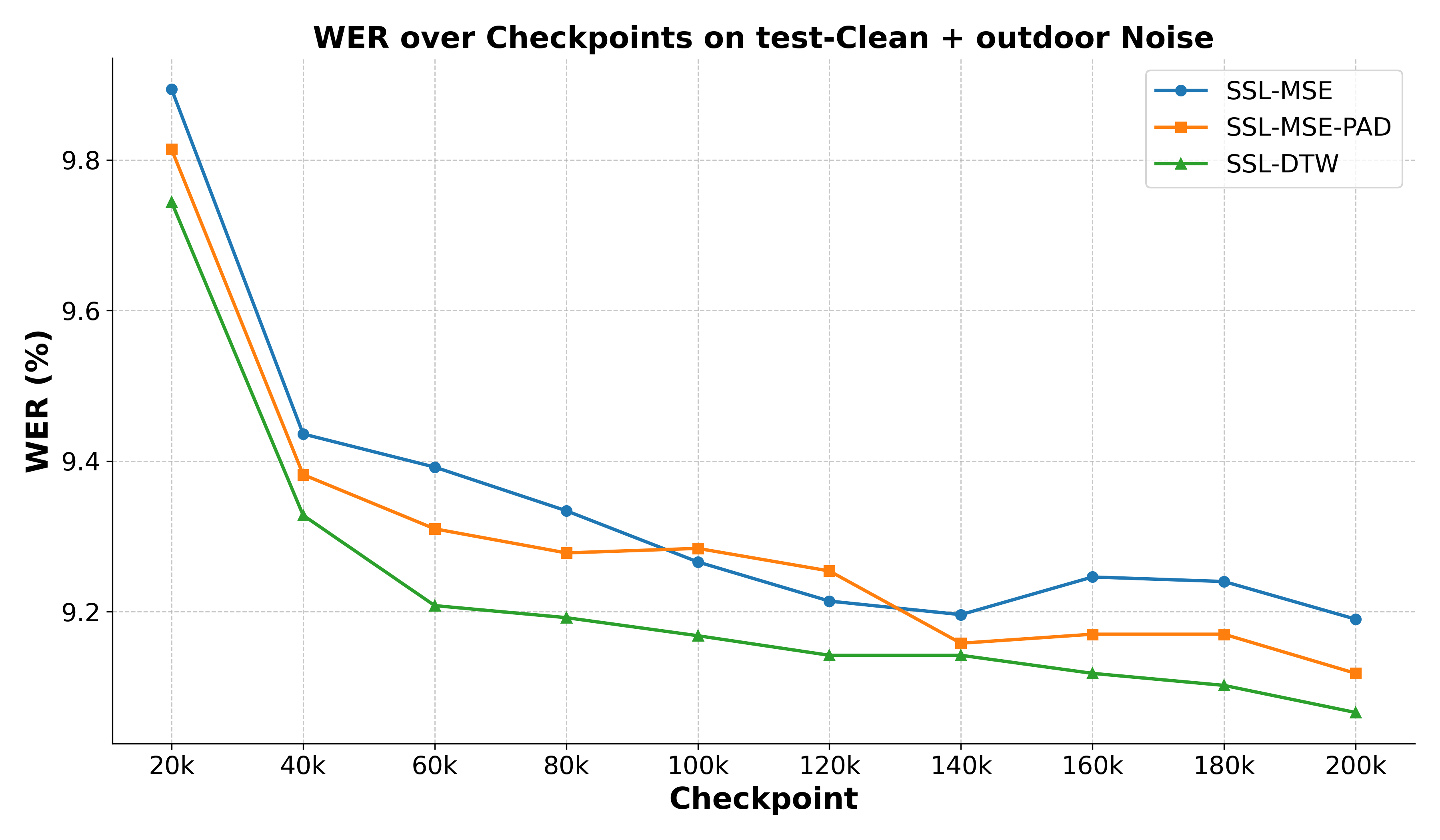}
    \caption{WER (in \%) on \texttt{test-clean + outdoor noise} for ASR across training checkpoints. SE frontends are fine-tuned with different objectives: SSL-MSE, SSL-MSE-PAD, and SSL-SoftDTW. Each curve shows the mean of 5 runs.}
    \label{fig2}
\end{figure}

Another key observation is convergence speed. Fig.~\ref{fig2} shows the WER on \texttt{test-clean + outdoor noise} across training checkpoints. SSL-SoftDTW converges significantly faster, reaching the final SSL-MSE performance in around 60k steps compared to 200k steps for SSL-MSE. SSL-MSE-PAD also exhibits faster convergence than SSL-MSE, although its final performance improvement is limited. Each curve represents the mean of five runs, with variance being negligible and omitted for clarity.  
It is noteworthy that addressing the exploitation of positional embeddings here occurs during SE fine-tuning, a relatively lightweight procedure, unlike in SPIRAL \cite{spiral} where positional issues arise during computationally expensive SSL pre-training. These findings suggest that incorporating such position-invariant strategies into pre-training itself may yield broader benefits across tasks and conditions.  

\paragraph*{Phoneme Recognition:}
Table \ref{tab3} reports results for the PR task under the same three test conditions. Similar to ASR, SSL-SoftDTW consistently improves robustness in unseen noise conditions compared to SSL-MSE. In contrast, SSL-MSE-PAD offers no improvement for PR. 
Taken together, the results in Tables \ref{tab1} and \ref{tab3} demonstrate that addressing the exploitation of positional embeddings through SSL-SoftDTW benefits downstream tasks by improving error rates and accelerating convergence. These findings highlight the value of position-invariant fine-tuning strategies in enhancing the robustness of SSL-guided SE models.

\section{Conclusions and Future Work}
\label{sec5}

This study examined a general issue associated with self-supervised learning (SSL) models, namely the exploitation of positional embeddings when mean squared error (MSE) is used as the fine-tuning objective. Two mitigation strategies were investigated: SSL-MSE-PAD and SSL-SoftDTW, using SSL representation-guided speech enhancement as a case study. The results showed that SSL-SoftDTW achieves faster convergence and improved performance on downstream tasks, whereas SSL-MSE-PAD, inspired by the SPIRAL framework \cite{spiral}, yielded only marginal improvements.  
An important observation is that these benefits were obtained by fine-tuning only the SE module, without modifying the SSL encoder or downstream models. This suggests that mitigating positional exploitation even at the level of a single component can provide measurable gains. Future work could explore applying these techniques during more computationally expensive stages such as SSL pre-training, where issues of positional dependence have previously been noted \cite{spiral}. Extending these methods to other scenarios where SSL-based losses are employed outside the pre-training setup also represents a promising direction, highlighting the broader importance of position-invariant fine-tuning for SSL-based speech modelling.  

\section{Acknowledgement}
This work was supported by the Centre for Doctoral Training in Speech and Language Technologies (SLT) and their Applications funded by UK Research and Innovation [grant number EP/S023062/1]. This work was also funded in part by LivePerson, Inc. We acknowledge IT Services at The University of Sheffield for the provision of services for High Performance Computing.

\bibliographystyle{IEEEbib}
\bibliography{strings,refs}

\end{document}